Tissue Classification and Whole-Slide Images Analysis via Modeling of the Tumor Microenvironment and Biological Pathways

Junzhuo Liu [a], Xuemei Du [b], Daniel Reisenbüchler [a], Ye Chen [b], Markus Eckstein [c,d,e,f], Christian Matek [c,d,e,f], Friedrich Feuerhake [g,h], Dorit Merhof [a,i]

[a] Faculty of Informatics and Data Science, University of Regensburg, Regensburg, Germany

[b] Department of Pathology, Beijing Friendship Hospital, Capital Medical University, Beijing, China

[c] Institute of Pathology, Universitätsklinikum Erlangen, Friedrich-Alexander-Universität Erlangen-Nürnberg (FAU), Erlangen, Germany

[d] Comprehensive Cancer Center Erlangen-EMN (CCC ER-EMN), Erlangen, Germany

[e] Comprehensive Cancer Center Alliance WERA (CCC WERA), Erlangen, Germany

[f] Bavarian Cancer Research Center (BZKF), Erlangen, Germany

[g] Institute for Pathology, Hanover Medical School, Hannover, Germany

[h] Institute for Neuropathology, University Clinic Freiburg, Freiburg i.Br., Germany

[i] Fraunhofer Institute for Digital Medicine MEVIS, Bremen, Germany

**Abstract**

Automatic integration of whole-slide images (WSIs) and gene expression profiles has demonstrated substantial potential in precision clinical diagnosis and cancer progression studies. However, most existing studies focus on individual gene sequences and slide-level classification tasks, with limited attention to spatial transcriptomics and patch-level applications. To address this limitation, we propose a multimodal network, BioMorphNet, which automatically integrates tissue morphological features and spatial gene expression to support tissue classification and differential gene analysis. For considering morphological features, BioMorphNet constructs a graph to model the relationships between target patches and their neighbors, and adjusts the response strength based on morphological and molecular-level similarity, to better characterize the tumor microenvironment. In terms of multimodal interactions, BioMorphNet derives clinical pathway features from spatial transcriptomic data based on a predefined pathway database, serving as a bridge between tissue morphology and gene expression. In addition, a novel learnable pathway module is designed to automatically simulate the biological pathway formation process, providing a complementary representation to existing clinical pathways. Compared with the latest morphology–gene multimodal methods, BioMorphNet's average classification metrics improve by 2.67%, 5.48%, and 6.29% for prostate cancer, colorectal cancer, and breast cancer datasets, respectively. BioMorphNet not only classifies tissue categories within WSIs accurately to support tumor localization, but also analyzes differential gene expression between tissue categories based on prediction confidence, contributing to the discovery of potential tumor biomarkers.

**1.Introduction**

Spatial transcriptomics [1], as an emerging gene expression profiling technology, has been widely applied to molecular-level analyses of cancer and the interpretation of biological functions [2-4]. Unlike traditional bulk RNA sequencing, which can only capture gene expression from mixtures of cells within specific regions of WSIs [5], spatial transcriptomics offers a key advantage by providing gene expression profiles with spatial resolution across WSIs. The spatially and transcriptionally heterogeneous expression of genes enables the investigation of intratumoral molecular heterogeneity [6] and spatial heterogeneity within tumor tissues [7, 8]. Although tissue morphology has long been utilized for tumor classification and grading [9, 10], it lacks information on molecular characteristics and biological mechanisms. Moreover, it is difficult to accurately classify certain tissue types solely based on morphological features. For instance, the Consensus Molecular Subtypes (CMS) classification system for colorectal cancer is driven by molecular signatures [11], rather than histopathological appearance. In this context, spatial transcriptomics serves not only as a valuable complement to traditional histological assessments but also provides deeper insights into the tumor microenvironment, enabling biologically informed classification and a better understanding of cancer progression.

Most existing AI-based methods for tissue classification and tumor localization adopt an implicit strategy, i.e. they do not directly predict the categories of tissue patches within WSIs. This is primarily because acquiring detailed annotations requires substantial time and expert knowledge. Instead, these methods typically perform slide-level classification and visualize the hidden layers of neural networks to generate heatmaps that indirectly localize tumor regions [12-14]. CLAM [12], as a representative example, is a weakly supervised framework based on multiple instance learning, which employs an attention mechanism to automatically identify instances within WSIs that are informative for classification. The heatmaps derived from attention scores indicate regions of interest that contribute to the current task. Although such implicit methods demonstrate potential in tumor localization, they are often limited to distinguishing between tumor and non-tumor regions, and lack the capacity for more precise category-level classification. In clinical practice, however, there are already established tumor grading systems tailored to specific cancer types, such as the Prostate Cancer Grading System [15] and the Nottingham Histologic Grade for breast cancer [16]. Furthermore, the training of these methods typically relies on a substantial scale of H&E-stained images with WSI-level annotations. While some unsupervised methods that do not require labels have been proposed [17, 18], their evaluation and accuracy remain unsatisfactory. For tissue classification methods based on spatial transcriptomics, most existing approaches can be categorized as unsupervised clustering algorithms driven by gene expression data [19-21]. These methods either rely solely on gene expression, or treat it as the dominant modality, with morphological features serving only as auxiliary information. However, such gene-driven clustering methods often focus on local regions and are limited in modeling or identifying distant but phenotypically similar tissue regions within WSIs. Another limitation is that pathologists, who are mainly trained to interpret morphological features, may find these methods less accessible, as their use typically involves gene filtering and clustering threshold tuning, which require specialized biological expertise and intuition.

Several recent studies have integrated tissue morphology and gene expression [22-26] for WSI-level classification. These methods typically employ two separate encoders to extract modality-specific features from images and gene expression profiles. Cross-attention mechanisms, positioned in the middle layers of the overall architecture, are then utilized to fuse multimodal features and generate modality-shared representations. Similar to morphology-only approaches, these multimodal fusion methods utilize hidden layer heatmaps to localize cancer regions. Although they incorporate gene expression data, such as bulk RNA-seq or copy number variation (CNV), these measurements are derived from partial WSI regions and lack spatial resolution. Consequently, they cannot establish a precise spatial correspondence between gene expression and tissue morphological features, limiting their ability to support more fine-grained tumor tissue classification and subsequent analyses of spatially differential gene expression.

Considering the limitations of unimodal methods and morphology–RNA fusion methods, integrating paired tissue morphological features with spatial gene expression for tissue classification is promising. Such an integrated approach facilitates automated tissue classification and tumor localization in clinical practice, and also links morphology with gene expression, enabling joint analysis of tumor morphological features and molecular profiles, contributing to improving tumor classification [27], understanding morphology changes from a molecular perspective [28], and analyzing tumor heterogeneity [29], among other applications.

In this work, we propose a novel multimodal fusion network that integrates tissue morphology and molecular biology to support tissue classification and molecular-level analysis of WSIs. First, in the morphological branch, our approach aims to represent and model the tumor microenvironment (TME), a complex and dynamic system composed of tumor cells, surrounding non-tumor cells, extracellular matrix, and signaling molecules. The TME plays a critical role in tumor initiation, progression, immune evasion, and therapeutic response [30]. Accordingly, we reflect the TME from two perspectives: tissue morphology and molecular expression. Specifically, for a target patch, not only its own morphological features but also those of its neighboring patches are incorporated into the representation of its histological context. To effectively model this microenvironment, we adopt a graph-based approach: the target patch is treated as the central node and connected to its spatial neighbors. Considering that tissues of the same type often appear in clusters, both morphological similarity and molecular expression similarity are computed to adjust the edge weights of the graph. This strategy captures the structural dependencies within the local context of the patch and, importantly, integrates the molecular functional state of the surrounding region through spatial transcriptomics. Second, the modality gap is explicitly considered. Since not all gene expression profiles are strongly associated with morphological features, directly fusing these modalities may interfere with effective model learning [31]. Inspired by biological encoding mechanisms and functional organization, pathway expressions are encoded from gene expression [32], and cross-attention is employed to integrate pathway expressions with tissue morphology. In this framework, pathways act as a semantic bridge connecting gene expression and morphological features. From the gene–pathway perspective, a pathway represents a biological process or functional module regulated by multiple genes. These genes are typically co-expressed or sequentially activated within regulatory networks, interacting with each other to precisely control specific cellular functions, physiological processes, or signal transduction events, thereby collectively constituting a complete biological pathway. From the pathway–morphology perspective, pathway-level signals more effectively capture tissue morphological states than individual gene expressions. Signaling pathways coordinate cellular behaviors and functions, ultimately shaping the morphological characteristics of tissues. Third, given that existing biological pathway databases do not comprehensively cover all known or potential pathways, and many remain under investigation without inclusion in curated knowledge databases, a learnable pathway sequence, which automatically models the process of encoding biological pathways from independent genes, is designed as a complement to clinically defined pathways. In addition, a gate layer is introduced at the end of the model to balance all contributing entities: tissue morphology, morphology–pathway fusion, and gene expression levels.

The main research contributions of this study are as follows:

1. A novel multimodal fusion network is proposed, which integrates tissue morphological features from WSIs and gene expression from spatial transcriptomics for tissue classification and WSI-level analysis.
2. The study provides insights into tissue classification and multimodal fusion from two perspectives. From the perspective of the tumor microenvironment, both local morphological context and molecular responses are considered when constructing the graph centered on the target patch. From the perspective of human biological function organization, pathway features, which more effectively reflect the morphological states, are encoded from spatial transcriptomic data. In addition to using prior pathway encodings from clinical biological pathway databases, a learnable pathway is introduced to simulate the gene-to-pathway formation process as a complement.

Furthermore, a multimodal fusion module is developed to integrate tissue morphology and pathway expression, enabling the construction of modality-shared features.

3. The proposed method achieves competitive performance, demonstrating strong potential for automatic tissue classification and tumor localization. Furthermore, prediction confidence is used as a threshold for automated differential gene expression analysis, facilitating downstream biological and pathological research, such as biomarker discovery.

**2. Methodology**

The general model design of our method is presented in Figure 1, comprising three branches: morphological feature encoding, spatial gene expression encoding, and morphology-pathway fusion. Paired patches and gene expression data serve as inputs, with the category prediction of the target patch as the output. In the morphological encoding branch, a graph is employed to model the tissue microenvironment, and morphological features are extracted using a graph convolutional neural network. In the morphology-pathway fusion branch, clinical pathway features are encoded from spatial transcriptomics data based on biological pathway databases. Additionally, a learnable pathway sequence is designed to simulate the gene-to-pathway formation process. This branch also incorporates a multimodal module to construct joint modality-shared features. At the end of the overall architecture, a gating layer and a classification head are used to adaptively weight the contributions of the three entities and generate the final prediction.

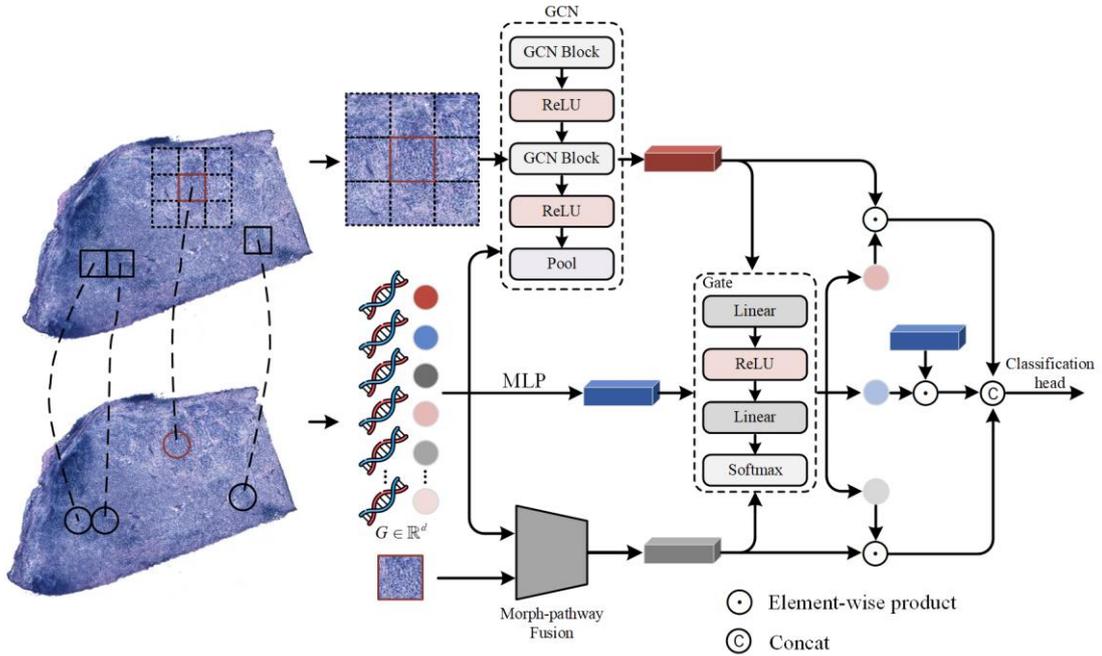

Fig. 1. The overall framework of BioMorphNet. It supports three input modalities: morphological features, biological pathway sequences, and spatial gene expression profiles.

A. Tissue Morphology Encoding

Tumorigenesis is not merely the result of autonomous behavior of individual cells, but rather the outcome of complex interactions between malignant cells and locally surrounding cells, tissues, and signaling molecules. Together, they form a dynamic ecosystem known as the TME [30]. Inspired by this concept, the tissue morphology encoding branch jointly models local tissue context from both morphological and spatial molecular perspectives, aiming to more accurately emulate the intricate organizational ecosystem of TME. Specifically, the histopathology foundation model UNI v1 [33] is employed to extract 1024-dimensional feature vectors from image patches and the graph is then constructed to model the local tissue context. For each target patch, eight adjacent patches are selected as neighbors. In the graph, the target patch serves as the central node connected to its adjacent patches (k = 8), forming an initial KNN graph that reflects the local tissue context. To more accurately capture the association strength between the target patch and its surrounding environment, we propose an edge weighting strategy based on similarity measurements. The TME modelling module is shown in Figure 2. From the morphological perspective, the mean squared error (MSE) between the image features of the central node and each of its eight neighbors is calculated, and the reciprocal of these scores is used as the morphological response. Likewise, from the molecular perspective, the MSE scores of gene expression between the central node and its neighbors are computed to obtain the molecular relationships. The average scores of the morphological and molecular relationships are used as the edge weights in the graph. The morphological encoding process can be formulated as follows:

$$r_i^{morph} = \frac{1}{MSE(x_c, x_i)}$$

$$r_i^{bio} = \frac{1}{MSE(g_c, g_i)}$$

$$e_i = \frac{r_i^{morph} + r_i^{bio}}{2}$$

where $r_i^{morph}$ denotes the morphological response, $r_i^{bio}$ denotes the molecular response, $e_i$ is the edge weight.

A graph convolutional neural network (GCN) is employed to extract morphological features from this graph structure. The architecture of the GCN is illustrated in Figure 1. The output dimension of the GCN is set to 512, serving as the final feature representation of the morphological encoding module. The morphological feature representation of the target patch is denoted as $x_c \in \mathbb{R}^{1024}$, and the features of its eight neighboring patches are represented as $\{x_i\}_{i=1}^{8}$. Their corresponding gene expression vectors are denoted as $g_c$ and $\{g_i\}_{i=1}^{8}$, respectively. This process can be formulated as follows:

$$h = GCN(x_c, \{x_i\}_{i=1}^{8}, \{e_i\}_{i=1}^{8})$$

where $h$ represents the output of the morphological encoding.

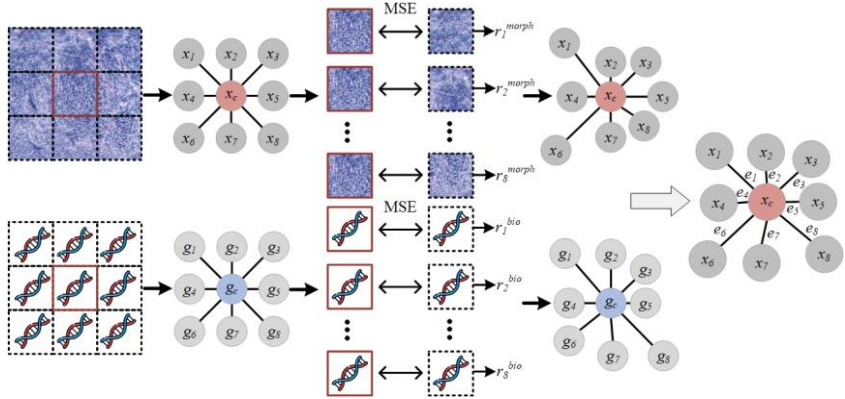

Figure 2. The workflow of graph construction for morphological encoding. Morphological features and gene expression profiles from the target patch and its neighboring patches are used as input. The resulting graph is subsequently fed into a GCN for feature extraction.

B. Morphology–Pathway Interaction

In cells, the synthesis of proteins is strictly regulated by the genetic information encoded in genes. Genetic alterations can thus directly affect the structure and function of proteins, thereby influencing key cellular behaviors such as proliferation, apoptosis, and invasion, often accompanied by adaptive morphological changes. Consequently, in whole-slide image analysis, the composition and variation of genes show a significant correlation with the resulting biological morphology of the corresponding cells. These functional changes are often orchestrated through biological pathways, which, from the biological perspective, are defined as series of interacting molecules or coordinated biological processes [34]. Functional alterations at the cellular level are frequently accompanied by morphological remodeling, and the cumulative effects across many cells ultimately manifest as observable morphological differences in tissue structure within the WSI [35]. Since tissue morphology is not directly shaped by individual genes, but rather emerges from the coordinated actions of gene sets influencing cellular behavior, biological pathways can be regarded as an intermediate modality that bridges gene expression and tissue morphology within the multimodal fusion branch of BioMorphNet. Figure 3 illustrates the morphology–pathway fusion module. Let the original spatial transcriptomic profile be given as $G \in \mathbb{R}^d$, where $d$ denotes the number of genes. Let $P = \{P_1, P_2, P_3, \cdots, P_m\}$ denote the collection of predefined biological pathways from the pathway database, where each $P_j \in \{1, \cdots, d\}$ represents the set of gene indices corresponding to the $j$-th pathway. Specifically, we compute the overlap score between the original spatial transcriptomic profile and each gene set associated with a biological pathway in the Reactome database [32]. A pathway is selected when its overlap score exceeds a predefined threshold (set to 0.9). The aggregated expression levels of all genes within a selected pathway are then used to represent the pathway activation level, serving as its molecular response signal. This process is formulated as follows:

$$Overlap(P_j) = \frac{|P_j \cap \{1,2,\cdots,d\}|}{|P_j|}$$

$$z_j = \sum_{i \in P_j} G_i$$

$$z = [z_1, z_2, \cdots, z_k] \in \mathbb{R}^k$$

Where $z_j$ denotes the expression level of the $j$-th pathway, and $z$ represents the final pathway-level feature vector.

Although curated pathway databases provide reliable biological priors, existing resources remain insufficient to capture the full complexity of biological processes. Many biological mechanisms have yet to reach consensus and require further experimental validation. Moreover, a vast number of intricate biological pathways remain unexplored by the scientific community. Due to these conditions, in addition to using predefined pathways, we introduce a novel learnable pathway sequence designed to model the process by which genes assemble into pathways, thereby serving as a complementary representation to clinical pathways. The learnable pathway sequence is illustrated in Figure 3(a). Specifically, we define a learnable weight matrix $w \in \mathbb{R}^{a \times d}$ to simulate the mapping from genes to $a$ learnable pathways ($a$ = 200). From a biological perspective, a single pathway may involve several to hundreds of genes. For each learnable pathway, the top 5% of genes are selected from the spatial transcriptomic profile consisting of thousands of genes, based on their corresponding learnable weights. The Softmax function is applied to the selected genes to assign normalized weights, reflecting their relative importance within the pathway. The final pathway response level is then obtained by computing a weighted sum of the original gene expression values, guided by the learned weights. Let the weight vector of the $i$-th pathway be $w_i \in \mathbb{R}^d$. The process is formulated as follows:

$$m_i = Top(w_i, 5\%)$$

$$a_i = Softmax(w_i \odot m_i)$$

$$z'_i = \sum_{j=1}^{d} a_{i,j} \cdot G_j$$

$$z' = [z'_1, z'_2, \cdots, z'_a] \in \mathbb{R}^a$$

where $Top(w_i, 5\%)$ denotes the operator that preserves only the largest 5% of the weights in wi and sets all remaining entries to zero. $m_i$ denotes the mask vector for the *i-th* learnable pathway, $a_i$ represents the gene-level weights after Softmax, $z'_i$ indicates the expression level of the *i-th* learnable pathway, and $z'$ denotes the final learnable pathway sequence. Two MLP are used to encode clinical pathways and learnable pathways. MLP consists of a linear layer, layer normalization, ReLU, and Dropout with a rate of 0.5. The output dimension of the linear layer is uniformly set to 512.

The fusion of morphological features and biological pathways is accomplished through Transformer modules equipped with a cross-attention mechanism. Specifically, the 1024-dimensional morphological feature vector of the target patch is first fed through an MLP to encode it to a 512-dimensional feature, aligning it with the dimensionality of the pathway vector. The architecture of the cross-attention mechanism is illustrated in Figure 3(c). It learns two separate attention matrices to facilitate interactions between the morphological features and the predefined pathways as well as the learnable pathways. For the predefined pathway vector, the morphological feature vector is linearly projected to generate the query (Q) and key (K), while the value (V) is derived from the predefined pathway. The attention matrix is computed by the dot product of Q and K, followed by a Softmax. This matrix is then multiplied with V to yield the fused representation $f_1$. A parallel cross-attention is performed between the morphological features and the learnable pathway vector, resulting in another fused feature $f_2$. To adaptively integrate these two branches, a gating mechanism is introduced. This module consists of two linear layers followed by a Softmax, which outputs two attention coefficients. $f_1$ and $f_2$ are concatenated along the feature dimension and fed into the gating mechanism. The resulting weights are then applied to $f_1$ and $f_2$, and the final fused feature is obtained through a weighted sum. This output is then passed into the MLP layer of the Transformer module. The overall morphology and pathway fusion module consist of two stacked Transformer blocks, each with 8 attention heads and a dropout rate of 0.25.

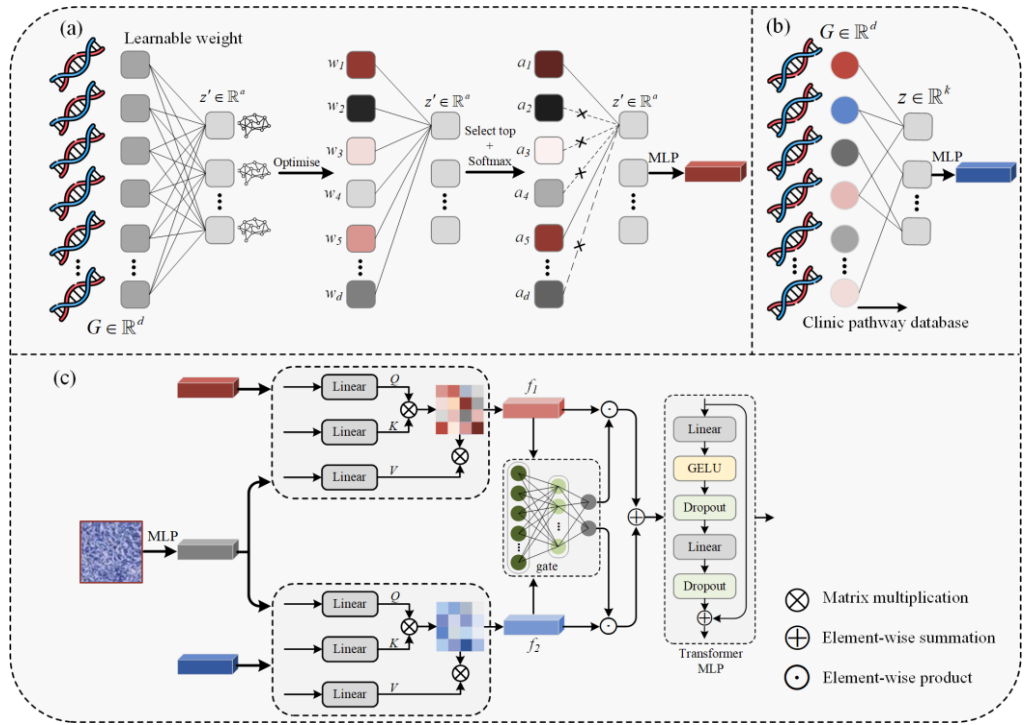

Fig. 3. Morph-Pathway Fusion Module. Spatial gene expression sequences are processed by: (a) A learnable pathway encoding, (b) A clinically informed pathway encoding. (c) Featurized patches are contextualized with processed pathway sequences with morphology-pathway fusion.

C. Spatial Transcriptomic Encoding

Certain individual genes, such as HER2 [36] in breast and MT1G [37] in prostate, have been recognized as tumor biomarkers. Although pathway-level encoding provides valuable insights from a molecular perspective, replacing individual gene expression entirely with pathway features may obscure the contributions of key biomarkers due to the presence of other genes within the same pathway. Therefore, in BioMorphNet, an independent branch for individual gene expression is preserved. This branch consists of a MLP comprising a linear transformation layer, layer normalization, ReLU activation, and a dropout layer with a rate of 0.5. The MLP maps the original gene expression vector into a 512-dimensional feature representation, which is then integrated into the late fusion module of BioMorphNet.

D. Loss Function

BioMorphNet employs a weighted cross-entropy loss function for classification, which encourages the model to pay more attention to minority classes and helps achieve a more balanced performance across different categories. The weight $W_i$ for the *i-th* category is computed as follows:

$$W_i = \frac{N}{C \cdot N_i}$$

where $N$ is the total number of samples, $C$ is the total number of classes, and $N_i$ denotes the number of samples in the *i-th* category. The weighted cross-entropy loss function is formulated as:

$$L = -\sum_{i=1}^{C} W_i \cdot y_i \cdot \log(\hat{y}_i)$$

where $y_i$ is the one-hot label and $\hat{y}_i$ is the predicted probability of the *i-th* category by the model.

3. Experiments and Results

A. Datasets and Evaluation Protocol

Three oncological datasets are used for evaluation, including prostate cancer [38], breast cancer [39], and colorectal cancer [6]. For each dataset, patch-level annotations are provided by clinical pathologists. The prostate cancer dataset contains 7 samples. Rare tissue types such as inflammation (38), fat (137), blood vessels (17), and patches that pathologists cannot classify are excluded (1,740), resulting in four tissue categories: stroma (8,623), benign (5,942), Gleason grading 4 cribriform (GG4; 4,575), and Gleason grading 2 (GG2; 1,416). The breast cancer dataset has 8 samples, including 6 tissue categories: cancer in situ (138), invasive cancer (1,757), connective tissue (616), adipose tissue (299), immune infiltrate (252) and breast glands (95). After filtering uncertain patches and aggregating categories, it consists of tumor (1,895) and normal tissue (1,262). The colorectal cancer dataset includes 6 samples with three tissue classes: tumor (3,662), stroma (4,787), and intestinal mucosa (2,124). The patches are normalized with mean values of [0.485, 0.456, 0.406] and standard deviations of [0.229, 0.224, 0.225]. For spatial transcriptomics data, Scanpy [19] is used to select common genes and filter out genes with zero expression. After preprocessing, the prostate cancer, colorectal cancer and breast cancer datasets contain 21,096, 22,580 and 12,674 genes, respectively. Evaluation metrics include Balanced Accuracy (Bal.ACC), Weighted F1-score (W-F1), AUPRC, and AUROC.

For the prostate cancer dataset, the samples are split as follows: 3 for training, 2 for validation, and 2 for testing. For the breast cancer dataset, the split is 4 for training, 2 for validation, and 2 for testing. For the colorectal cancer dataset, the split is 2 for training, 2 for validation, and 2 for testing. This process is repeated randomly five times, and the average results are computed. PyTorch serves as the deep learning framework, and all experiments run on a server equipped with an Nvidia RTX A5000 24G GPU. We employed the AdamW optimizer and trained for a maximum of 60 epochs with batch size 32, weight decay 1e-4 and learning rate 1e-4.

B. Competing Methods

To the best of our knowledge, no existing work addresses patch-level classification based on both histopathology and spatial transcriptomics. Therefore, to benchmark our proposed method, we selected several competitive multimodal architectures which are designed to integrate visual morphology with gene expression data, including PathomicFusion [22], MCAT [24], CMTA [40], MOTCAT [25], MIF [41], HealNet [42]. These methods have demonstrated advanced performance on WSI classification tasks incorporating histological morphology and gene expression data (e.g., RNA-Seq, SNV). All methods are trained and evaluated under the same settings.

C. Classification Results

Table 1 demonstrates the classification performance of BioMorphNet and competitive methods, indicating that BioMorphNet achieves significant improvements across all datasets. The superior Balanced Accuracy and Weighted F1 demonstrate that BioMorphNet achieves higher classification accuracy, particularly under class-imbalanced scenarios. The higher AUPRC and AUROC further indicate stronger discriminatory power, enabling the model to reliably distinguish between different tissue types. Overall, BioMorphNet exhibits robust generalization across cancer types and provides more effective tissue classification capability in comparison to other methods.

Table 1. Classification results on prostate cancer, colorectal cancer and breast cancer datasets.

| Datasets | Models | Bal.ACC | W-F1 | AUPRC | AUROC | Mean |
|---|---|---|---|---|---|---|
| Prostate | PathomicFusion | 0.743±0.098 | 0.767±0.087 | 0.855±0.073 | 0.931±0.033 | 0.824±0.067 |
| | MCAT | 0.636±0.116 | 0.709±0.085 | 0.810±0.092 | 0.886±0.046 | 0.760±0.072 |
| | CMTA | 0.754±0.073 | 0.797±0.065 | 0.893±0.051 | 0.948±0.025 | 0.848±0.049 |
| | MOTCAT | 0.678±0.038 | 0.718±0.045 | 0.811±0.062 | 0.899±0.039 | 0.777±0.039 |
| | MIF | 0.766±0.034 | 0.814±0.064 | 0.903±0.057 | 0.955±0.024 | 0.860±0.040 |
| | HealNet | 0.772±0.055 | 0.813±0.070 | 0.890±0.053 | 0.952±0.026 | 0.857±0.049 |
| | **BioMorphNet** | **0.801±0.074** | **0.829±0.087** | **0.931±0.049** | **0.970±0.023** | **0.883±0.052** |
| Colorectal | PathomicFusion | 0.710±0.038 | 0.651±0.109 | 0.808±0.028 | 0.882±0.016 | 0.763±0.042 |
| | MCAT | 0.521±0.179 | 0.447±0.222 | 0.659±0.148 | 0.740±0.119 | 0.592±0.165 |
| | CMTA | 0.638±0.040 | 0.554±0.079 | 0.762±0.104 | 0.840±0.074 | 0.699±0.059 |
| | MOTCAT | 0.516±0.173 | 0.443±0.219 | 0.643±0.167 | 0.723±0.134 | 0.581±0.172 |
| | MIF | 0.728±0.093 | 0.685±0.105 | 0.839±0.064 | 0.884±0.051 | 0.784±0.076 |
| | HealNet | 0.675±0.054 | 0.602±0.128 | 0.842±0.058 | 0.889±0.047 | 0.752±0.049 |
| | **BioMorphNet** | **0.752±0.078** | **0.718±0.089** | **0.905±0.033** | **0.932±0.024** | **0.827±0.047** |
| Breast | PathomicFusion | 0.756±0.063 | 0.762±0.039 | 0.900±0.022 | 0.854±0.063 | 0.818±0.042 |
| | MCAT | 0.687±0.113 | 0.668±0.133 | 0.897±0.038 | 0.867±0.028 | 0.780±0.074 |
| | CMTA | 0.710±0.122 | 0.740±0.092 | 0.918±0.037 | 0.893±0.053 | 0.815±0.064 |
| | MOTCAT | 0.692±0.093 | 0.665±0.103 | 0.898±0.042 | 0.867±0.035 | 0.781±0.063 |
| | MIF | 0.755±0.076 | 0.769±0.053 | 0.909±0.023 | 0.873±0.047 | 0.827±0.038 |
| | HealNet | 0.745±0.085 | 0.773±0.040 | 0.921±0.025 | 0.889±0.034 | 0.832±0.035 |
| | **BioMorphNet** | **0.819±0.046** | **0.837±0.008** | **0.942±0.016** | **0.918±0.019** | **0.879±0.016** |

Figure 4 presents the confusion matrices of multiple models, illustrating their classification performance across different tissue categories. Overall, BioMorphNet demonstrates superior classification capability in most categories. Notably, it correctly identifies 90.4% samples in the GG4 category, outperforming other competing methods such as CMTA (83.0%) and MIF (85.2%). In addition, BioMorphNet achieves the highest number of correctly classified samples in the stroma category (89.6%), significantly reducing misclassification. One limitation of BioMorphNet lies in its performance on the GG2 category, where 40.6% GG2 patches are misclassified as stroma. On the colorectal cancer dataset, BioMorphNet correctly identifies 66.3% tumor tissues (second only to MIF) and 81.0% intestinal mucosa tissues (second only to PathomicFusion). A particular challenge arises in distinguishing between stroma and tumor regions, where 21.3% tumor tissue patches are misclassified as stroma, and 24.6% stromal regions are incorrectly predicted as tumor. Figure 4 (c) shows the confusion matrices for the breast cancer dataset. BioMorphNet exhibits higher sensitivity to tumor tissues, correctly identifying 86.7% tumor samples, outperforming the competing models. For normal tissues, BioMorphNet correctly identifies 80.6% samples, which is less than MCAT (86.9%) and MOTCAT (90.3%). Nevertheless, MCAT misclassifies 40.4% tumor samples as normal tissue, and MOTCAT even misclassifies 45.4%, revealing a pronounced imbalance in their predictions.

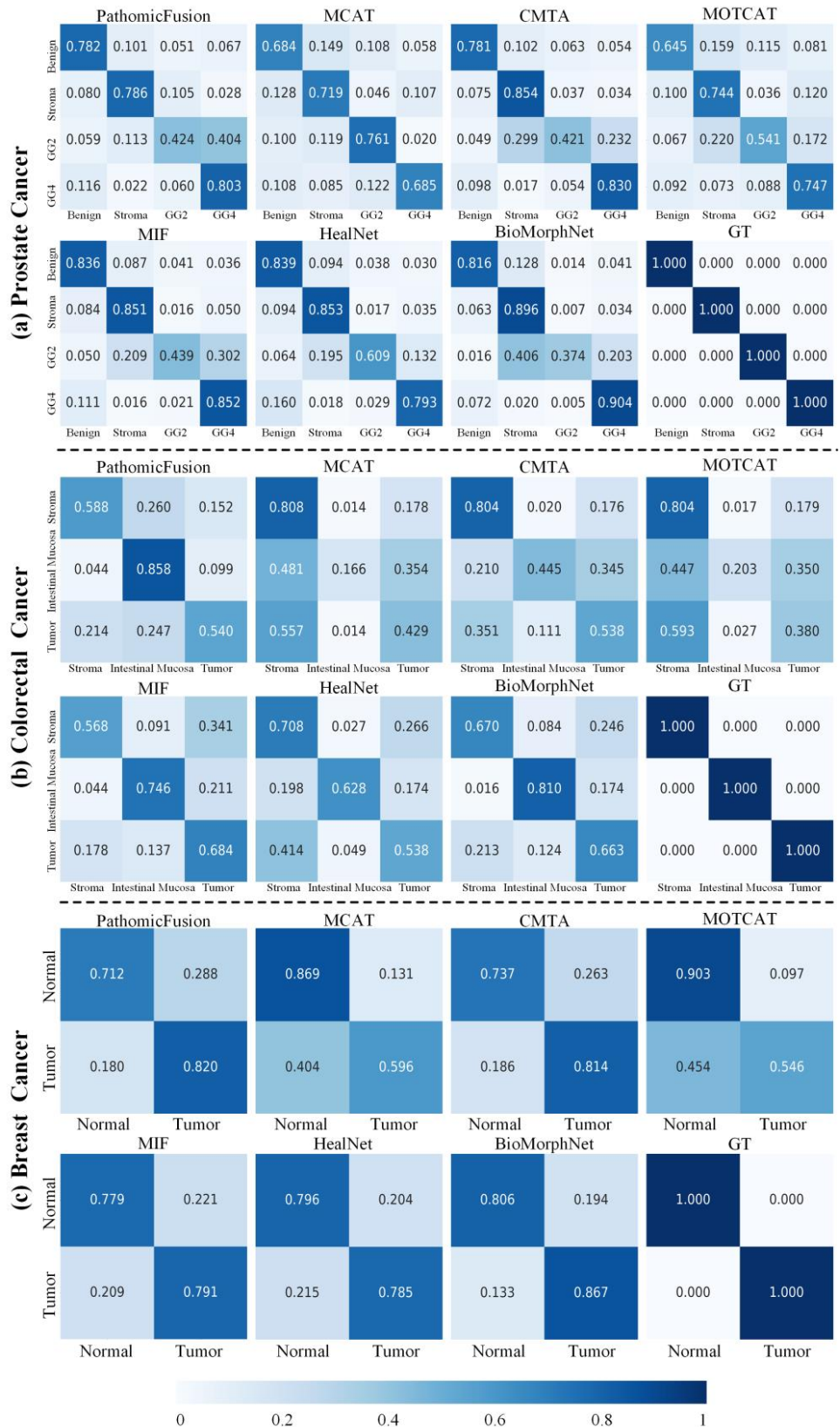

Figure 4. Confusion matrices on three datasets. The horizontal axis represents the predicted classes by the model, and the vertical axis represents the ground truth labels. (a) Prostate cancer dataset. (b) Colorectal cancer dataset. (c) Breast cancer dataset.

D. Visualization of Tissue Category Predictions

Figure 5 illustrates the classification predictions for a prostate cancer sample containing three tissue types: benign, stroma, and GG2. In this sample, only a small portion of GG2 tissue is available in the lower right corner of the WSI, while most of the tissue consists of benign and stroma. PathomicFusion, CMTA, and MOTCAT incorrectly classify some regions as GG4. MCAT fails to identify GG2 and instead misclassifies it as stroma. In contrast, MIF, HealNet, and BioMorphNet successfully identify GG2, which appears as three distinct clusters. However, both MIF and HealNet misclassify some benign regions as GG2, as shown in the last row of Figure 5.

Figure 6 presents the classification predictions for a colorectal cancer sample. BioMorphNet achieves an overall accuracy of 0.885, outperforming other competitive methods. For tumor tissue, BioMorphNet better preserves the spatial integrity of tumor predictions, highlighting the importance of effectively modeling the tumor microenvironment.

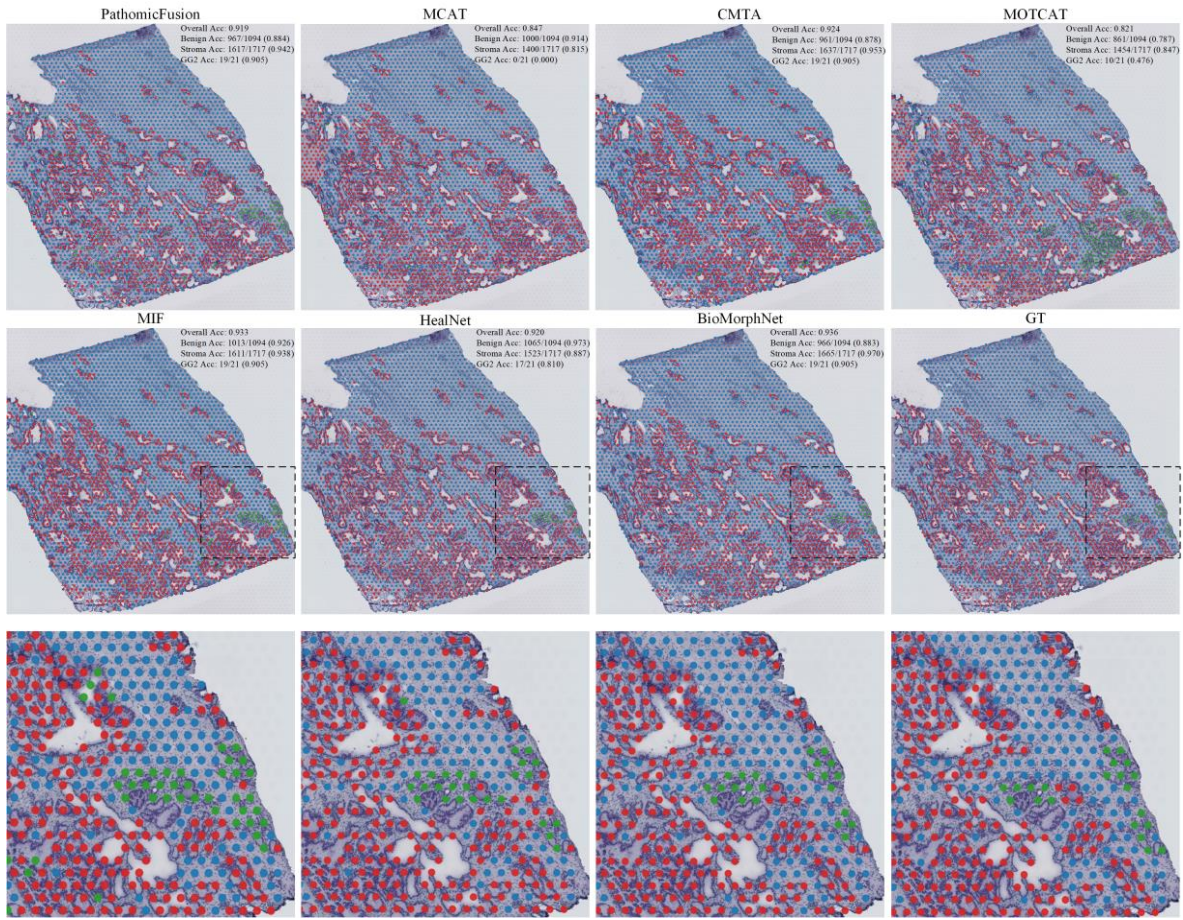

Figure 5. Visualization of classification predictions on the prostate cancer dataset. Red: Benign. Blue: Stroma. Green: GG2. Orange: GG4.

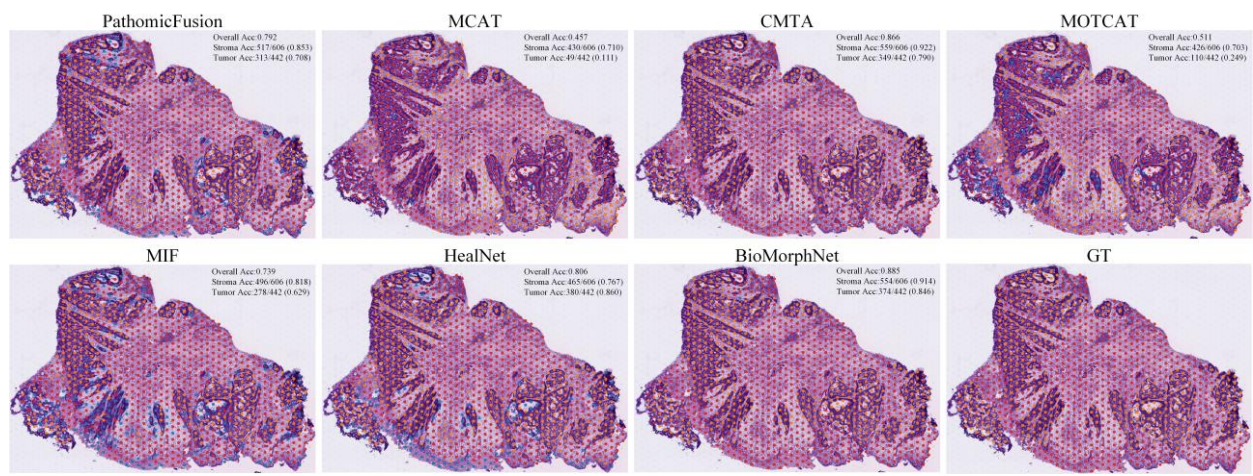

Figure 6. Visualization of classification predictions on the colorectal cancer dataset. Red: Stroma. Orange: Tumor. Blue: Intestinal mucosa.

E. Biologically Informed Analysis of WSIs

To evaluate BioMorphNet's potential in identifying cancer-related genes, Scanpy is employed to perform automated differential gene expression analysis on the test samples. Patches with high-confidence predictions (confidence score ≥ 0.95) are collected from the test set. In the prostate cancer dataset, a total of 11,714 patches are gathered, including stroma (5,654), benign (2,533), GG2 (172), and GG4 (3,355). In the breast cancer dataset, 1,408 patches are collected, including 990 tumor and 418 normal tissues. In the colorectal cancer dataset, 4,383 patches are collected, including stroma (1,991), tumor (1,672) and intestinal mucosa (720). The Wilcoxon rank-sum test is employed to support differential gene expression analysis across tissue categories. Figure 7(a) presents the results for the prostate cancer dataset. Notably, PPFIA2 is highly expressed in GG4 tissue while exhibiting near-zero expression in other tissues, showing a clear statistical significance. This finding is consistent with recent studies [43], suggesting that PPFIA2 is associated with Gleason score upgrading in prostate cancer and may serve as a potential biomarker for studying disease progression. For the benign tissue, MT1G shows significant differential expression, with its expression markedly downregulated in tumor and stromal tissues. This observation is supported by clinical studies [37], indicating that MT1G overexpression inhibits prostate cancer cell growth both in vitro and in vivo and is associated with poor prognosis. Figure 7(b) displays the differential expression results for the breast cancer dataset. Several clinically recognized breast cancer marker genes are identified, including DDX5 [44], CD24 [45], and ERBB2 (HER2) [36]. These genes play critical roles in breast cancer development. For instance, DDX5 is involved in cell proliferation and transcriptional regulation; CD24 is linked to cell adhesion and metastasis; and ERBB2 (HER2) amplification is a key molecular hallmark of HER2-positive breast cancer, commonly used to guide targeted therapy. Figure 7(c) presents the differentially expressed genes predicted by BioMorphNet in colorectal cancer. Several markers that are markedly downregulated in tumor tissues, such as ITLN1 [46] and PLA2G2A [47], are successfully detected.

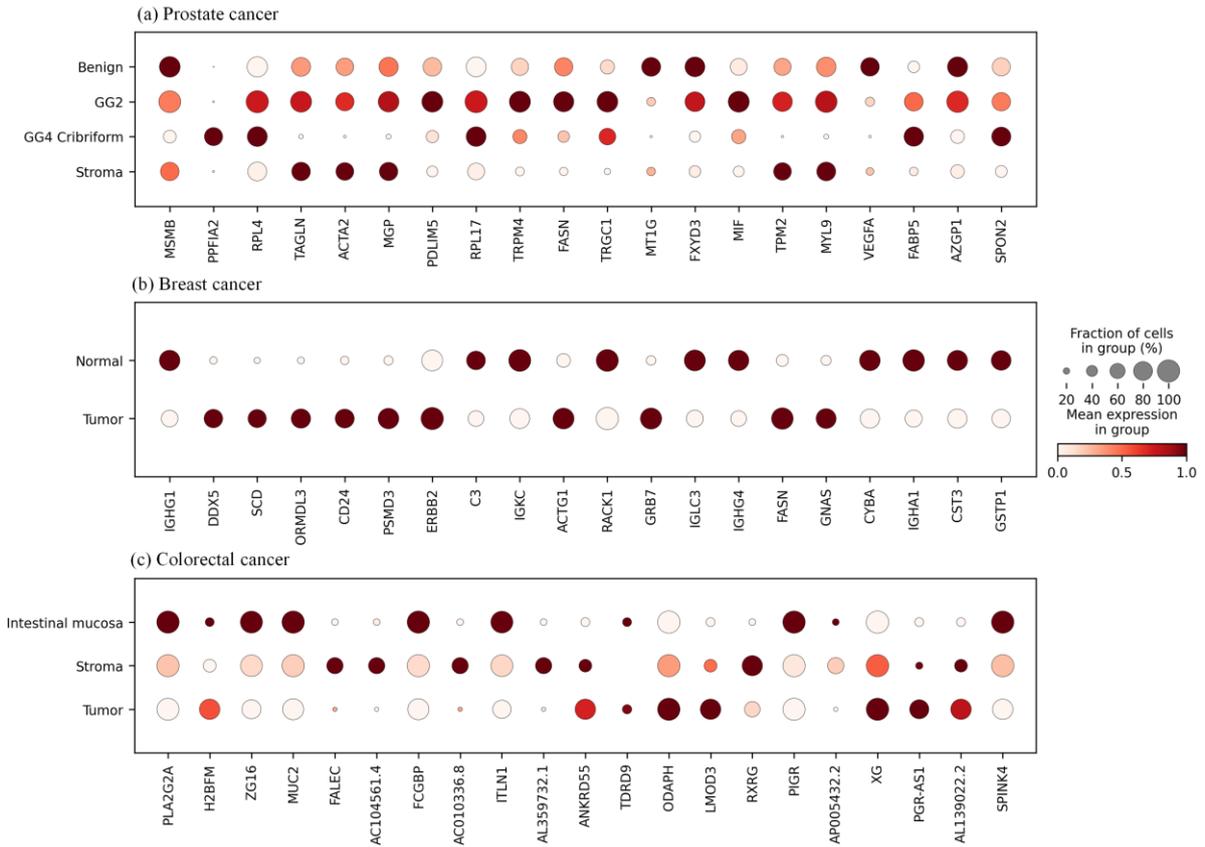

Figure 7. Differential gene expression analysis results across tissue categories. All identified differentially expressed genes are significant with p < 0.001, as inferred by a Wilcoxon rank-sum test. The x-axis shows the differentially expressed genes, and the y-axis represents the tissue categories. The size of each dot indicates the proportion of patches within the corresponding tissue category that express the gene. The color denotes the average (normalized) expression level of the gene. (a) Prostate cancer dataset. (b) Breast cancer dataset. (c) Colorectal cancer dataset.

F. Ablation Study

Table 2 presents the ablation study results. Both morphological features and molecular sequences exhibit reasonable classification capability. The application of concatenation and late fusion strategies leads to slight performance improvements, as shown in the third row. Introducing the graph to model the contextual information around the target patch results in a notable performance gain. The improvements introduced by clinically predefined biological pathways and learnable pathways are substantial in both the colorectal cancer and breast cancer datasets. When the two types of pathways are integrated, the model exhibits enhanced classification performance across all three datasets.

Figure 8 presents the impact of the number of learnable pathways on model performance. We observe that the influence of the number of learnable pathways on classification performance varies across datasets. In the prostate and breast cancer datasets, increasing the number of learnable pathways leads to relatively stable performance, ranging from 0.869–0.887 for prostate cancer and 0.864–0.879 for breast cancer. In contrast, adding more learnable pathways does not benefit the classification of colorectal cancer tissues. These differences likely reflect variations in gene expression diversity and underlying biological mechanisms across cancer types, resulting in distinct response patterns to pathway capacity. Overall, a moderate number of learnable pathways (such as 100 or 200) tends to yield more robust tissue classification, while larger pathway sets do not provide substantial performance gains.

Table 2. Ablation study results on three datasets. The average metric is reported. All experiments are performed with the same experimental equipment and parameter settings.

| Image | Pathway | ST | Prostate | Colorectal | Breast |
|---|---|---|---|---|---|
| Seq. | × | × | 0.837 | 0.770 | 0.830 |
| × | × | √ | 0.765 | 0.607 | 0.737 |
| Seq. | × | √ | 0.845 | 0.772 | 0.837 |
| Graph | × | √ | 0.874 | 0.806 | 0.852 |
| Graph | Clinic | √ | 0.876 | 0.822 | 0.862 |
| Graph | Learnable | √ | 0.875 | 0.825 | 0.862 |
| Graph | Clinic + Learnable | √ | 0.883 | 0.827 | 0.879 |

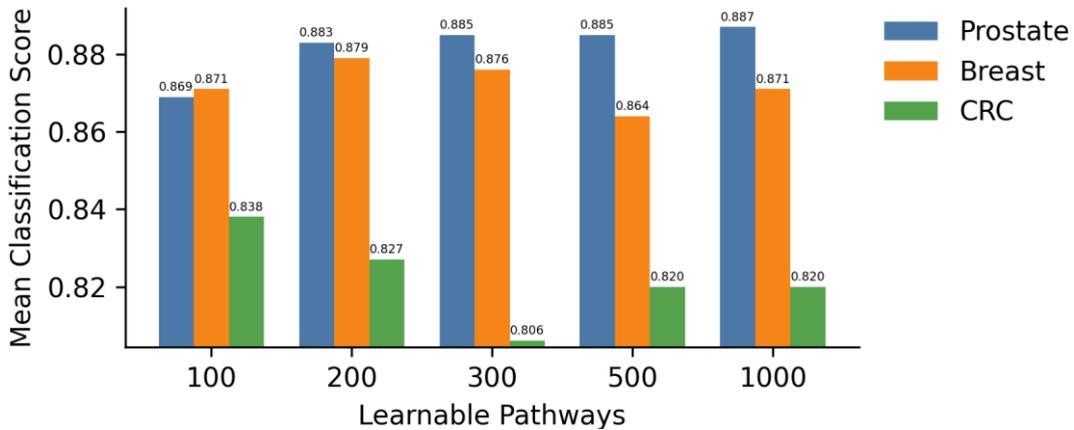

Figure 8. Hyperparameter study of learnable pathway number. The x-axis indicates the number of learnable pathways, and the y-axis represents the average of all classification metrics.

## 4. Discussion

The interaction between tissue morphology and gene expression demonstrates significant potential in the automatic diagnosis of cancer. However, most current studies mainly focus on WSI-level classification and the integration of WSI-level RNA-Seq or CNV data. These molecular profiles represent average expression across the WSIs, lacking insight into internal spatial heterogeneity. Although some studies [12-14] generate heatmaps using patch-level attention scores to localize tumor regions, such localizations are often coarse and typically limited to binary classification between tumor and normal tissue, which falls short of the fine-grained grading required in clinical settings (e.g., Gleason grading in prostate cancer). Some studies attempt to refine suspicious patches using clustering algorithms, but the resulting clusters still require manual interpretation by clinical experts. In contrast, BioMorphNet is a spatial transcriptomics-based framework tailored for patch-level tasks, offering a more detailed integration of tissue morphology and molecular context in cancer research. It focuses on intra-slide tissue structures and expression patterns, enabling the direct prediction of more fine-grained tissue categories and supporting accurate tumor localization. Moreover, BioMorphNet provides a more realistic molecular expression context, as gene expression can be precisely mapped to specific WSI regions based on transcript coordinates. By combining predicted labels with spatially resolved differential gene expression analysis, clinicians and clinical researchers can investigate expression differences across regions of the tumor microenvironment, identifying regulatory patterns between tumor cores and margins, which is crucial for understanding cancer progression, metastatic potential, and drug response mechanisms.

Modeling the TME leads to a significant improvement of performance. We observe that tissue regions of the same category tend to form local clusters rather than appearing in a scattered manner. Relying solely on the target patch may result in suboptimal performance, particularly when the morphological features of the patch are inconspicuous, such as those located at the tumor boundary. In clinical practice, TME analysis is also widely used to support cancer progression studies, therapeutic response prediction, and prognosis assessment. Although existing methods [48] incorporate contextual information of patches, they often fail to account for the interaction strength between the central patch and its surrounding regions. In contrast, BioMorphNet not only generates differential visual responses based on morphological similarity, but also integrates spatial transcriptomic information to capture biologically meaningful responses from a molecular perspective.

For the integration of molecular expression and morphology, several studies [49] also encode biological pathways from gene expression and subsequently fuse them with morphological representations. These studies typically begin by selecting relevant pathways based on clinical pathway databases, and then encode the expression profiles of genes within each pathway into fixed-dimensional tokens. In contrast, BioMorphNet does not assign a separate token to each pathway. Instead, it directly computes the aggregated expression level of all genes within a given pathway. This design is motivated by two considerations. First, assigning a token to every pathway substantially increases model complexity and parameter count. Second, since the number of genes varies across pathways, enforcing a fixed-dimensional token for each pathway may result in inconsistent information density across tokens. Another limitation of these methods is that the biological pathways are constructed based on WSI-level averaged gene expression, which leads to the loss of spatial variability. In fact, tumor and normal regions within a WSI often exhibit significant functional differences, and the activation level of the same pathway can vary significantly across regions. Moreover, the learnable pathway module as a complement to clinically defined pathways. It enhances the model's flexibility and scalability in biological representation and shows potential for further improving the performance of multi-modal fusion.

One limitation of BioMorphNet is its reliance on the scale of training data, a common challenge for many supervised learning methods. The classification results indicate suboptimal performance on the GG2 subtype of prostate cancer tissue. We observe that most misclassified GG2 patches originate from a single WSI containing a large GG2 region. This suggests that, when GG2 samples are underrepresented in the training set, BioMorphNet tends to misclassify them as stroma. Nevertheless, as shown in Figure 5, the model achieves significantly better performance on GG2 when sufficient training samples are available. Another limitation lies in the sparsity of spatial transcriptomics data. We observe that within the breast cancer dataset, each WSI contains several hundred transcription points, meaning these points cannot cover the entire tissue, potentially introducing bias into the evaluation. By contrast, spatial transcriptomics data for prostate and colorectal cancers exhibit superior quality. Due to the scarcity of WSIs with tissue labels and spatial transcriptomics data, the breast cancer dataset is solely used for evaluation. Nevertheless, higher-quality datasets must be incorporated into future research.

In clinical practice, BioMorphNet could contribute to pathological diagnosis and biological gene analysis. First, it can automatically identify tissue types on WSIs, enabling rapid pre-screening and automated annotation of key regions to support pathologists in clinical diagnosis. Second, by combining tissue type predictions with spatial transcriptomics for automated differential gene expression analysis, BioMorphNet can uncover spatially resolved molecular features associated with distinct tissue types, thereby providing biological evidence for prognostic assessment and the identification of potential therapeutic targets. It provides comparison groups for differential gene expression analysis that are both histologically well-defined and spatially contiguous, thereby alleviating the subjectivity associated with relying solely on unsupervised clustering or manually delineated regions of interest (ROIs). Overall, BioMorphNet integrates morphological and molecular information, thereby directly connecting histopathological cancer diagnosis with molecular-level profiling.

# 6.Conclusion

This study proposes a novel architecture, BioMorphNet, that integrates tissue morphology with spatial gene expression for patch-level classification within WSIs. Its design is inspired by two aspects: (1) clinical practices involving tumor microenvironment (TME) analysis, and (2) the molecular regulatory mechanisms by which independent genes interact through signaling pathways to modulate cellular function and ultimately influence morphological phenotypes. BioMorphNet models the TME by constructing the graph that captures both the contextual information of target patches and their interaction with neighboring regions. In addition, it incorporates a learnable biological pathway sequence module to complement predefined clinical pathways and simulate gene-to-pathway regulatory processes. Experimental results demonstrate that it not only accurately identifies tumor regions in WSIs but also shows potential for composite biomarker discovery derived from interactions between visual tissue morphology and molecular biology, contributing to the understanding of interactions between morphology and molecular biology. Future work will focus on creating patch-level annotations for more datasets.